\pdfoutput=1

\documentclass[11pt]{article}

\usepackage[preprint]{acl}

\usepackage{times}
\usepackage{latexsym}

\usepackage[T1]{fontenc}

\usepackage[utf8]{inputenc}

\usepackage{microtype}

\usepackage{inconsolata}

\usepackage{inconsolata}
\usepackage{graphicx}
\usepackage{multirow}
\usepackage{subcaption}
\usepackage{caption}
\usepackage{booktabs}
\usepackage{tcolorbox}
\usepackage{enumitem}
\usepackage{amsmath}
\usepackage{makecell}
\usepackage{graphicx}
\usepackage{array}
\usepackage{amsmath}
\usepackage{amsmath}
\usepackage{amsthm}
\usepackage{booktabs}
\usepackage{algorithm}
\usepackage{algorithmic}

\title{Large Language Models are Limited in Out-of-Context Knowledge Reasoning}
\author{
Peng Hu\textsuperscript{$\clubsuit$}, Changjiang Gao\textsuperscript{$\clubsuit$}, Ruiqi Gao\textsuperscript{$\clubsuit$}, Jiajun Chen\textsuperscript{$\clubsuit$}, Shujian Huang\textsuperscript{$\clubsuit$} \\
\textsuperscript{$\clubsuit$}National Key Laboratory for Novel Software Technology, Nanjing University \\
\texttt{\{hup, gaocj, 201220184\}@smail.nju.edu.cn, \{chenjj, huangsj\}@nju.edu.cn}
}

\begin{document}
\maketitle
\begin{abstract}
Large Language Models (LLMs) possess extensive knowledge and strong capabilities in performing in-context reasoning. However, previous work challenges their out-of-context reasoning ability, i.e., the ability to infer information from their training data, instead of from the context or prompt. This paper focuses on a significant aspect of out-of-context reasoning: Out-of-Context Knowledge Reasoning (OCKR), which is to combine multiple knowledge to infer new knowledge. We designed a synthetic dataset with seven representative OCKR tasks to systematically assess the OCKR capabilities of LLMs. Using this dataset, we evaluated several LLMs and discovered that their proficiency in this aspect is limited, regardless of whether the knowledge is trained in a separate or adjacent training settings. Moreover, training the model to reason with reasoning examples does not result in significant improvement, while training the model to perform explicit knowledge retrieval helps for retrieving attribute knowledge but not the relation knowledge, indicating that the model's limited OCKR capabilities are due to difficulties in knowledge retrieval. Furthermore, we treat cross-lingual knowledge transfer as a distinct form of OCKR, and evaluate this ability. Our results show that the evaluated model also exhibits limited ability in transferring knowledge across languages. \footnote{Our code and data is available at: \url{https://github.com/NJUNLP/ID-OCKR}.}

\end{abstract}

\section{Introduction}
In the realm of in-context learning, LLMs not only demonstrate significant reasoning capabilities~\citep{kojima2022large, yao2023tree, besta2023graph} but also concurrently exhibit expertise as knowledge bases in various academic and professional domains, including science, history, law, and finance~\citep{petroni2019language, wei2023kicgpt, alkhamissi2022review}. However, it is unclear whether their reasoning ability is limited to in-context scenarios, or they can also perform out-of-context reasoning, which, as defined by previous studies \cite{berglund2023taken}, is ``to recall facts learned in training and use them at test time, despite these facts not being directly related to the test-time prompt.'' Berglund et al. \shortcite{berglund2023taken} showed that LLMs adapt their responding behaviors based on the given identity and the information about the identity in the training corpus. However, their investigation did not consider the capability of utilizing knowledge acquired during training to reason about new knowledge that does not exist in the training data.

\begin{figure}[h]
    \centering
    \includegraphics[width=0.48\textwidth]{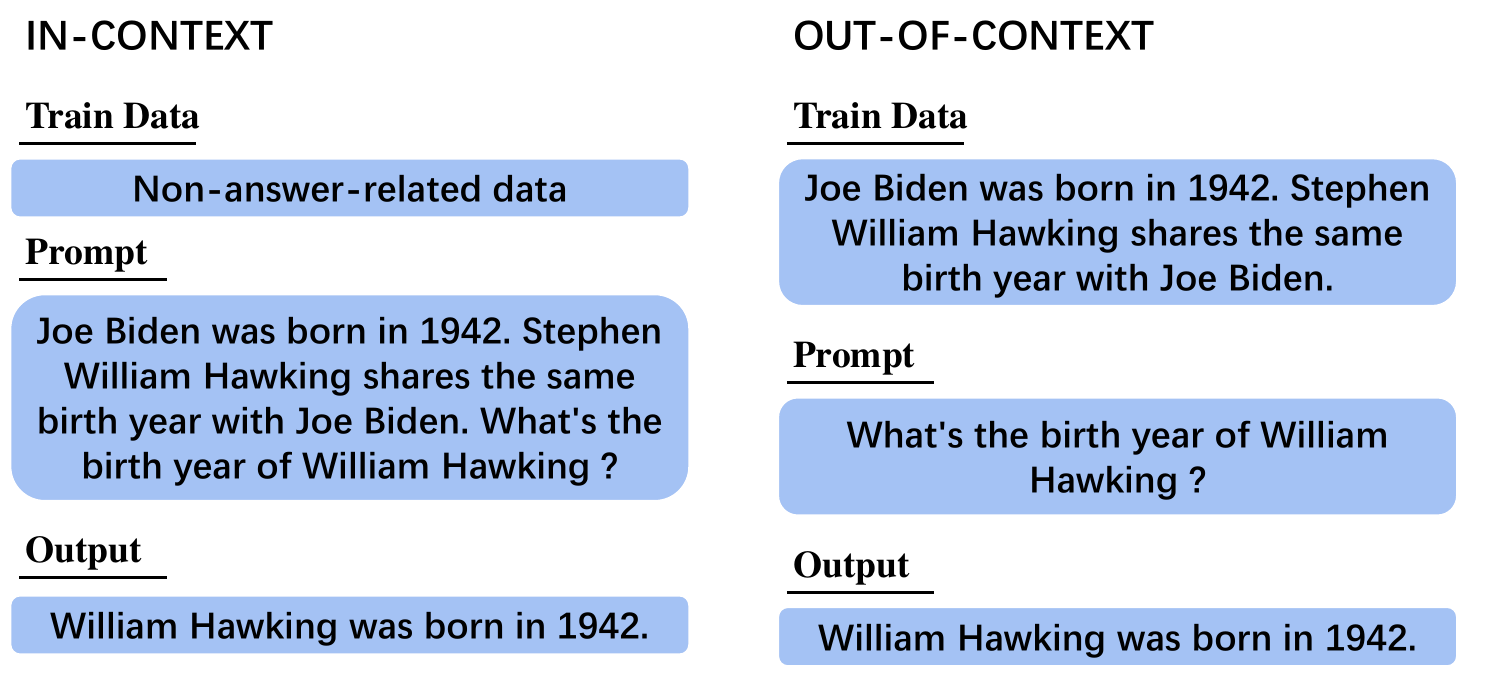}
    \caption{In-Context vs Out-of-Context. In the In-Context scenario, the relevant data is provided in the prompt to allow the model to infer the answer. In the Out-of-Context scenario, the relevant data is included directly in the training data, and the model is then asked to infer the answer based on this training.}
    \label{fig:INCONTEXT}
\end{figure}

For instance, if an LLM knows from the training data that \textit{Joe Biden was born in 1942} and \textit{Stephen William Hawking shares the same birth year with Joe Biden}, can it infer \textit{Hawking's birth year} as 1942 without having been directly trained on this specific fact? This kind of reasoning can be more intuitively understood by being compared with In-Context Learning (ICL) (Figure \ref{fig:INCONTEXT}). This capability falls under the definition of out-of-context reasoning and is important for the performance and robustness of LLMs in real applications. 

\begin{table*}[ht]
\footnotesize
\centering
\begin{tabular}{lll}
\toprule
\textbf{Combination} & \textbf{Examples of \(T_{1}\) and \(T_{2}\)} & \textbf{Feasibility and possible \(\bar{T}\)} \\
\midrule
$A \land A\rightarrow A$ & 
\begin{tabular}[c]{@{}l@{}}
(x, birth\_year, 2000)\\ 
(y, birth\_year, 2000)
\end{tabular} & No, cannot infer new meaningful attributes \\\hline

$A \land A\rightarrow R$ & 
\begin{tabular}[c]{@{}l@{}}
(x, birth\_year, 2000)\\ 
(y, birth\_year, 2000)
\end{tabular} & Yes, e.g.: (x, birth\_year\_equals, y) \\\hline

$A \land R\rightarrow A$ ($R \land A\rightarrow A$) & 
\begin{tabular}[c]{@{}l@{}}
(x, birth\_year, 2000)\\ 
(x, birth\_year\_equals, y)
\end{tabular} & Yes, e.g.: (y, birth\_year, 2000) \\\hline

$A \land R\rightarrow R$ ($R \land A\rightarrow R$) & 
\begin{tabular}[c]{@{}l@{}}
(x, birth\_year, 2000)\\ 
(x, birth\_year\_equals, y)
\end{tabular} & No, cannot infer new meaningful relationships \\ \hline

$R \land R\rightarrow A$ & 
\begin{tabular}[c]{@{}l@{}}
(x, birth\_year\_equals, y)\\ 
(x, birth\_year\_equals, z)
\end{tabular} & No, pure relationships cannot infer attributes \\ \hline

$R \land R\rightarrow R$ & 
\begin{tabular}[c]{@{}l@{}}
(x, birth\_year\_equals, y)\\ 
(x, birth\_year\_equals, z)
\end{tabular} & Yes, e.g.: (y, birth\_year\_equals, z) \\ 
\bottomrule
\end{tabular}
\caption{Feasibility analysis of all possible combinations for the reasoning patterns. x, y, and z denote specific entities involved in the training process. Considering the interchangeability of \(T_{1}\) and \(T_{2}\), redundant combinations are eliminated. For $A \land A\rightarrow A$ and $A \land R\rightarrow R$, it is difficult to derive meaningful new knowledge without borrowing other external knowledge. For $R \land R\rightarrow A$, attributes cannot be inferred from pure relationships. Consequently, we identify $A \land A\rightarrow R$, $A \land R\rightarrow A$, and $R \land R\rightarrow R$ as viable knowledge reasoning patterns.}
\label{tab:combination}
\end{table*}

This paper proposes the investigation of Out-of-Context Knowledge Reasoning (OCKR), a vital component of out-of-context reasoning. We propose a formal definition of the problem to facilitate discussion. We discuss and design 7 related tasks covering reasoning over different kind of knowledge, such as attributes (A) and relations (R), and construct corresponding datasets to systematically evaluate the OCKR abilities. 
The evaluation on several open-source LLMs, e.g. LLaMA2-13B-CHAT \cite{touvron2023llama}, Baichuan2-13B-CHAT \cite{yang2023baichuan}, Pythia-12B \cite{biderman2023pythia}, LLaMA3-8B-Instruct \cite{touvron2023llama}, shows that these LLMs have very limited OCKR ability.

Intuitively, new knowledge can emerge during the training or inference phase. We also conduct experiments to assist the LLMs to perform OCKR in different phases, which serve as in-depth analyses for the potential difficulties of reasoning.
In the training phase, we merge related knowledge into adjacent text, which may be easier for reasoning.
In the inference phase, we train the LLMs to learn the reasoning pattern, or provide them with chain-of-thought (COT) prompt, explicitly retrieving and applying the knowledge.
We also study the cross-lingual OCKR as a special case.





Our main findings are:

\begin{itemize}
\item All the evaluated models show limited OCKR ability, no matter the required knowledge occurs adjacently or separately during training. In comparison, the reasoning could be easily done in a in-context setting. 
\item Training the model with reasoning examples does not lead to significant improvement, suggesting that the reasoning ability in general might not be the bottleneck for OCKR.
\item Training the model with explicit retrieval steps help the model achieve higher accuracy in one task (retrieving attribute knowledge and inferring a relation). However, when tasked with retrieving relational knowledge, the performance stays at random levels. This suggests that besides the model's limitation in automatically performing knowledge retrieval, it faces significant challenges in accurately retrieving relational knowledge, which limits its efficacy in OCKR.
\item Cross-lingual performance surpasses the random level, indicating that learning the translation relation may be different from learning other monolingual relations. There are diversities among languages, whilethe overall performance remains weak.
\end{itemize}

\section{Problem Definition}

\subsection{OCKR Problems}
An example of OCKR can be formally represented as:
\begin{equation}
    T_{1} \land T_{2} \land \ldots \land T_{n} \rightarrow \bar{T} \quad (n \geq 1)
\end{equation}
where \( T_{1},T_{2},\ldots,T_{n} \) denotes knowledge in training data; \( \bar{T} \) denotes knowledge not in the training data; with the constraint that \( T_{1},T_{2},\ldots,T_{n}\) are sufficient to imply  \( \bar{T} \). If a given model trained on \( T_{1},T_{2},\ldots,T_{n} \)  can correctly answer question about  \( \bar{T} \), , we say that the model has $n$-ary OCKR ability, i.e. the model can infer \( \bar{T} \) from \( T_{1},T_{2},\ldots,T_{n} \).

In this paper, we focus on the binary OCKR case where n = 2, i.e., \( T_{1} \land T_{2} \rightarrow \bar{T}\), which is the simplest case that allows knowledge to be reasoned between different entities.

The knowledge considered in this study falls into two categories according to the knowledge graph taxonomy: Attributes (A) and Relations (R). They are involved with entities in triplets, i.e., (Entity, Attribute, Value) for attributes, (Entity, Relation, Entity) for relations \cite{kejriwal2021knowledge}. 

By using A and R as the known knowledge and infer new knowledge, six potential combinations can be enumerated. Among them, only three combinations can be aligned with feasible knowledge reasoning patterns. They are: Attribute $\land$ Attribute $\rightarrow$ Relationship ($A\land A\rightarrow R$), Attribute $\land$ Relationship $\rightarrow$ Attribute ($A \land R\rightarrow A$), and Relationship $\land$ Relationship $\rightarrow$ Relationship ($R \land R\rightarrow R$). See Table \ref{tab:combination} for details. Thus, we choose these three types of reason tasks for further study.



\begin{figure*}[ht]
\centering
\includegraphics[width=0.75\linewidth]{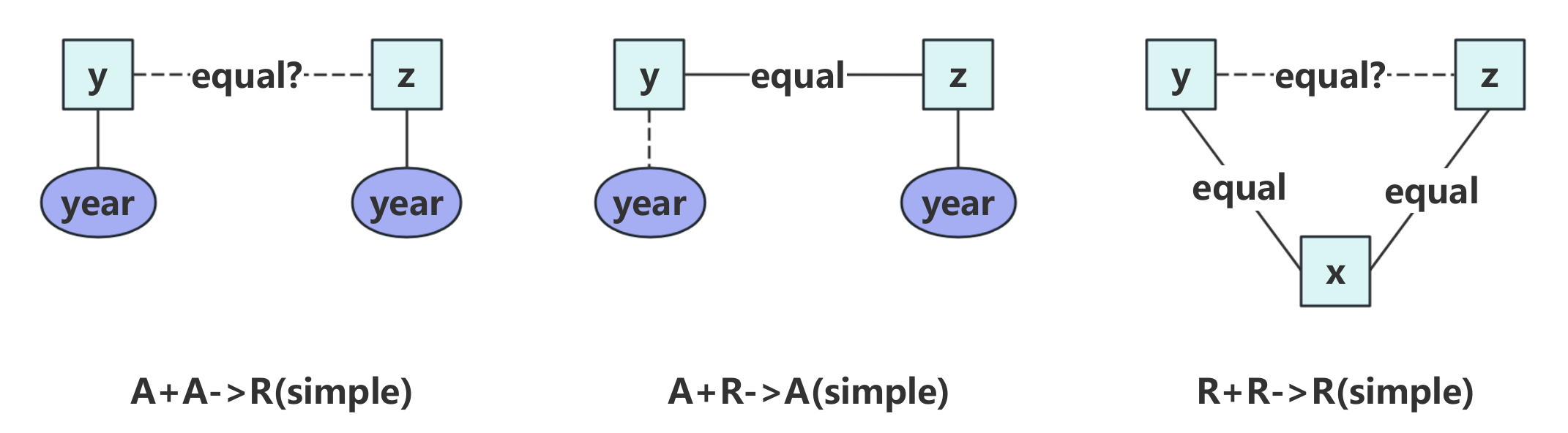}
\caption{The diagram shows the entities, attributes and relations in the dataset, for simple versions of the three reasoning patterns. Rectangles denote entities, ellipses indicate attributes, and edges represent relationships. Solid black lines represent knowledge in the training data, while dashed black lines represent knowledge in the test data. As the reasoning examples (Sec. \ref{sec:reasoningTraining}), a portion of the knowledge represented by the dashed black lines are provided to the training process for learning the corresponding inference patterns. The model is then tested on the knowledge represented by the remaining dashed black lines.}

\label{fig:complete-chain-reasoning}
\end{figure*}

\begin{table*}[ht]
\footnotesize
\centering
\begin{tabular}{p{0.2\linewidth}p{0.35\linewidth}p{0.35\linewidth}}
\toprule
\textbf{Reasoning Patterns} & \textbf{Knowledge Templates of Training Data} & \textbf{Knowledge Template of Test Data} \\
\midrule
$A \land A\rightarrow R$ (simple) & 
\begin{tabular}[c]{@{}l@{}}
(y, birth\_year, year)\\ 
(z, birth\_year, year)
\end{tabular} & 
\begin{tabular}[c]{@{}l@{}}
(y, birth\_year\_equals, z)\\ 
(y, not\_birth\_year\_equals, z$'$)
\end{tabular} \\

\hline

$A \land R\rightarrow A$ (simple) & 
\begin{tabular}[c]{@{}l@{}}
(x, birth\_year, year)\\ 
(x, birth\_year\_equals, y)\\ 
(x$'$, not\_birth\_year\_equals, y)
\end{tabular} & 
(y, birth\_year, year) \\

\hline
$R \land R\rightarrow R$ (simple) & 
\begin{tabular}[c]{@{}l@{}}
(x, birth\_year\_equals, y)\\ 
(x, not\_birth\_year\_equals, y$'$)\\ 
(x, birth\_year\_equals, z)\\ 
(x, not\_birth\_year\_equals, z$'$)
\end{tabular} & 
\begin{tabular}[c]{@{}l@{}}
(y, birth\_year\_equals, z)\\ 
(y, not\_birth\_year\_equals, z$'$)
\end{tabular} \\

\hline
$A \land A\rightarrow R$ (hard) & 
\begin{tabular}[c]{@{}l@{}}
(y, birth\_year, year)\\ 
(z, birth\_year, year)
\end{tabular} & 
\begin{tabular}[c]{@{}l@{}}
(y, birth\_year\_greater\_than, z)\\ 
or (y, not\_birth\_year\_greater\_than, z)
\end{tabular} \\
\hline
$A \land R\rightarrow A$ (hard) & 
\begin{tabular}[c]{@{}l@{}}
(z, birth\_year, year)\\ 
(y, one\_year\_older\_than, z)\\ 
(y, not\_one\_year\_older\_than, z$'$)
\end{tabular} & 
(y, birth\_year, year $+1$) \\
\hline
$R \land R\rightarrow R$ (hard) & 
\begin{tabular}[c]{@{}l@{}}
(x, birth\_year\_greater\_than, y)\\ 
(x, not\_birth\_year\_greater\_than, y$'$)\\ 
(y, birth\_year\_greater\_than, z)\\ 
(y$'$, not\_birth\_year\_greater\_than, z$'$)
\end{tabular} & 
\begin{tabular}[c]{@{}l@{}}
(x, birth\_year\_greater\_than, z)\\ 
(x, not\_birth\_year\_greater\_than, z$'$)
\end{tabular} \\
\hline
Cross-lingual & 
\begin{tabular}[c]{@{}l@{}}
(x$_{\text{en}}$, birth\_year, year)\\ 
(x$_{\text{en}}$, translation, x$_{\text{L}}$)\\ 
\end{tabular} & 
(x$_{\text{L}}$, birth\_year, year) \\
\bottomrule
\end{tabular}
\caption{Overview of ID-OCKR: This table summarizes the reasoning patterns together with the data templates included in the dataset. The prime symbol ($'$) in y$'$ distinguishes it from y. The subscript L in x$_{\text{L}}$ stands for languages other than English. z$_{\text{older}}$ represents z that is older than or equal to y's birth year, and z$_{\text{younger}}$ represents z that is younger than y's birth year.}
\label{tab:datasets}
\end{table*}

\subsection{Dataset Design}

This paper introduces the Inference Dataset for OCKR (ID-OCKR). The dataset encompasses seven subsets, including the three knowledge reasoning patterns, each presented at both simple and hard levels, along with a subset specifically designed for evaluating cross-lingual capabilities. See Table \ref{tab:datasets} for details.

\paragraph{Knowledge} Assessing the model's OCKR capabilities is non-trivial, because it is not easy to discriminate whether the knowledge is derived from the training data or actually exists in the training data. Furthermore, the LLMs language ability has a huge impact on its performance in different benchmarks. 
Therefore, it is essential to create a fictional set of knowledge that doesn't rely on knowledge of existing facts, and minimize the language barrier in understanding the knowledge. 

Therefore, we choose a very simple attribute, i.e. the year of birth, and some simple relations based on this single attribute, i.e. birth in the same year, birth year greater (i.e. older), one year older, etc, to avoid complex knowledge understanding. For adding a little challenge in the reasoning process, there are two levels of tasks. For the simple level of the task, the relation is only about the equivalence of the attributes; while for the hard level, the relation may need a numerical comparison or calculation of attributes. See Figure \ref{fig:complete-chain-reasoning} for an illustration of the three simple reasoning patterns.


\paragraph{Cross-lingual Task} The motivation for constructing a cross-lingual dataset stems from our recognition of translation as a unique relation type that links an entity to its translated counterpart. This allows us to conceptualize cross-lingual knowledge transfer as involving three components: attribute knowledge in English ($A$), translation knowledge (relation between English entity and the corresponding entity in another language, i.e. $R$), and attribute knowledge the other language (denoted as $A$). Thus, the cross-lingual scenario can be formally represented as a special form of $A \land R\rightarrow A$. 

As English is the dominant language in most LLMs, we mainly consider the knowledge transfer from English to other languages.
To capture a wide range of linguistic diversity, we selected nine languages based on their widespread use and diverse linguistic families: German (de), French (fr), Italian (it), Russian (ru), Polish (pl), Arabic (ar), Hebrew (he), Chinese (zh), and Japanese (ja). See Table \ref{tab:language_countries} for more details.

\begin{table}[ht]
    \centering
    \footnotesize
    \begin{tabular}{ccc}
\toprule
\textbf{ISO} & \textbf{Language} &\textbf{Language Family} \\ 
\midrule
en & English &Germanic \\
de & German &Germanic \\ \hline
fr & French &Romance \\
it & Italian &Romance \\ \hline
pl & Polish &Slavic \\
ru & Russian &Slavic \\ \hline
ar & Arabic &Afro-Asiatic \\
he & Hebrew &Afro-Asiatic \\ \hline
ja & Japanese &Japonic \\ \hline
zh & Chinese &Sino-Tibetan \\ 
\bottomrule
\end{tabular}
    \caption{Correspondence between Languages and Language Families}
    \label{tab:language_countries}
\end{table}

\subsection{Datasets Construction}
We utilize GPT-4~\cite{achiam2023gpt} to create fictitious entities and templates for our dataset. The dataset is then created based on these templates, entities, and predefined rules.

For entities, the names are constructed using fantastical words (e.g., ``ReverentDawn'') to ensure they are rare in the original corpus. For templates, the knowledge templates are generated based on the knowledge types described in Table \ref{tab:datasets}. We generate 10 text templates for each knowledge type to ensure the model can adequately capture the relevant knowledge. For the test set, only one text template is used to evaluate model performance.

For predefined rules, the generated entities are randomly assigned attributes (e.g., random birth years between 1991 and 2010). The relations between entities are determined based on these attributes. Examples of the generated instances are presented in Table \ref{tab:knowledge_triples} in Appendix \ref{sec:DataSample}. For additional details on the organization of datasets, please refer to Appendix \ref{sec:detailDataset}.

\begin{table*}[!t]
\centering
\footnotesize
\begin{tabular}{p{0.18\linewidth}p{0.72\linewidth}}
\toprule
\textbf{Reasoning Pattern} & \textbf{Retrieval Training Example} \\
\midrule
$A \land A\rightarrow R$ \newline (simple) & 
Q: Did BlissfulHarbor and FrostMirage share the same birth year? \textit{Please analyze the birth years of BlissfulHarbor and FrostMirage before giving your answer}. \newline
A: We know that BlissfulHarbor was born in the year 2010, and FrostMirage was born in the year 2010. Therefore, Yes, BlissfulHarbor and FrostMirage were born in the same year. \\
\hline
$A \land R\rightarrow A$ \newline (simple) & 
Q: In what year was XenoGlimmer born? \textit{Use the birth year of another person, who was born in the same year as XenoGlimmer, as a reference point to deduce the answer indirectly}. \newline
A: We know that XenoGlimmer and BlissfulHarbor were born in the same year, and BlissfulHarbor was born in the year 2010. Therefore, XenoGlimmer was born in the year 2010. \\
\hline
$R \land R\rightarrow R$ \newline (simple) & 
Q: Did XenoGlimmer and MeadowGlitter share the same birth year? \textit{Use the birth year of another person, who was born in the same year as XenoGlimmer, as a reference point to deduce the answer indirectly.} \newline
A: We know that XenoGlimmer and BlissfulHarbor were born in the same year, and BlissfulHarbor and MeadowGlitter were born in the same year. Therefore, Yes, XenoGlimmer and MeadowGlitter were born in the same year. \\
\bottomrule
\end{tabular}
\caption{Retrieval Training Examples for Three Simple Reasoning Patterns:  The italian font identifies the instruction to retrieve specific knowledge, which is appended to the question. The answer in the training examples are also augmented with the corresponding retrieval and reasoning steps. The questions and answers are generated utilizing connection templates generated by GPT-4.}
\label{tab:cot_examples}
\end{table*}

\section{Methodology}

\subsection{Evaluation of OCKR}
We perform training and test using the ID-OCKR dataset. For training, we fine-tune the LLMs so that the accuracy of responses to knowledge triples in the training set exceeds 90\% to ensure the LLM learns the knowledge.

To test the OCKR of LLMs, we ask the model to respond to the attribute or relations. 
For the assessment of attributes, we use an exact matching of values. Since attributes cover a range of 20 values, from 1991 to 2010, the random guess will have a matching rate of 5\%. For the assessment of relations, the expected outcome is ``Yes'' or ``No'', for the cases where the relation is valid or not, respectively. So the random level of matching rate is 50\%.
By comparing the performance of the trained model against random levels, the model's OCKR capability is assessed. 

Intuitively, the inference of new knowledge is a complex process involving both knowledge retrieval and reasoning. It is also possible that the LLMs are inferring new knowledge during either training or inference phase. Simply evaluating the overall performance does provide more insights into the detailed abilities.
To better understand the process of OCKR, we also carried out evaluation and analyses in the following scenarios where the LLMs are assisted in different ways.

\subsection{Assisting OCKR with Adjacent Knowledge}
 In real-world training, different knowledge are separated in different parts of the training data, which may make it hard to perform direct inference with them. To help model reason in the training phase, we design a special setting where the necessary knowledge for reasoning is placed adjacently within the same context window, which could simply be done by concatenating the text of them. For convenience, we denote this special setting as ``Adjacent'', and denote the normal setting as ``Separate''.



\subsection{Assisting OCKR with Reasoning Training}\label{sec:reasoningTraining}

Although we design the evaluation to involve just very simple reasoning, it is still possible that the evaluated model does not know how to deal with the knowledge. Thus we train the model with examples of reasoning, which illustrate the required type of reasoning, aiming to enhance the model's reasoning capabilities. 

More specifically, in case the model does not recognize that the type of knowledge of \(T_{1}\) and \(T_{2}\) infer \(\bar{T}\), we incorporate a number of (\(T_{1}\), \(T_{2}\), \(\bar{T}\)) as examples into the training set. If the model can understand the reasoning pattern in these examples, it may be able to reason for other cases where \(\bar{T}\) not explicitly present in the training data. To elaborate, we introduced additional data into the simple versions of the three knowledge reasoning patterns, as illustrated in Figure \ref{fig:complete-chain-reasoning}. 


\begin{table*}[!ht]
\footnotesize
\centering
\begin{tabular}{lrrrr}
\toprule
\textbf{Reasoning Pattern} & \textbf{Random} & \textbf{Separate} & \textbf{Adjacent} & \textbf{In-Context} \\
\midrule
$A \land A\rightarrow R$ (simple) & 50.0 & 50.8 & 51.8 & 100.0 \\
\addlinespace
$A \land R\rightarrow A$ (simple) & 5.0 & 5.0 & 6.0 & 100.0\\
\addlinespace
$R \land R\rightarrow R$ (simple) & 50.0 & 50.5 & 52.5 &  89.3\\
\addlinespace
$A \land A\rightarrow R$ (hard) & 50.0 & 50.8 & 52.6 & 84.7\\
\addlinespace
$A \land R\rightarrow A$ (hard) & 5.0 & 4.0 & 6.0 & 100.0 \\
\addlinespace
$R \land R\rightarrow R$ (hard) & 50.0 & 52.3 & 51.5 & 86.5 \\
\bottomrule
\end{tabular}
\caption{Performance Comparison Across Datasets and Scenarios: This table compares model performance across various datasets and experimental settings. 
}
\label{tab:mainResult}
\end{table*}

\subsection{Assisting OCKR with Retrieval Training}
Training with reasoning examples may help the reasoning, but does not explicitly teach the model how to perform OCKR. Idealy, the model may need to retrieve existing knowledge first, then perform reasoning with them.
Insprired by the CoT approach \cite{kojima2022large}, we explicitly lead the model to perform the retrieval and reasoning step, which further assesses the model's knowledge retrieval capabilities. 

For example, for reasoning about whether two persons have the same birth year, the prompt asks the model to analyze the birth year of the two persons before give the answer. More examples are shown in Table \ref{tab:cot_examples}.

To make sure the model correctly apply the CoT reasoning, we trained the model for each reasoning pattern with specific question and answer pairs, which carries the step-by-step reasoning (Table~\ref{tab:cot_examples}). With this assistance, the model learns to perform the required knowledge retrieval before reasoning \cite{ho2022large}.

\subsection{Evaluation of Cross-Lingual OCKR}
We evaluate the cross-lingual OCKR task, where the only relationship considered is the translation relation. 
The evaluation is performed in both the Separate and Adjacent training settings. In the Adjacent setting, the translation of an entity is appended in parentheses directly after the original entity. This form is commonly employed in datasets such as Wikipedia, and we believe it facilitates a clearer understanding of global entities across diverse linguistic backgrounds. See Table \ref{tab:knowledge_triples} in Appendix~\ref{sec:DataSample} for examples .





\section{Experiments}
\subsection{Experiment Setup}
The evaluation primarily utilized the LLaMA2-13B-CHAT model, trained using the Low-Rank Adaptation (LoRA) approach \cite{hu2021lora}. We also trained Baichuan2-13B-CHAT and Pythia-12B models with LoRA, and trained LLaMA2-7B-CHAT and LLaMA3-8B-Instruct\cite{touvron2023llama} models with full-finetune, as a supplement to the main experiment. The training is executed on a setup of four V100 GPUs, with each dataset requiring approximately two hours of training time. The experimental parameters and additional details can be found in Appendix \ref{sec:Hyper-parameters}.

\subsection{Basic OCKR results}
We conduct evaluation on six datasets, with both the standard ``Separate'' and ``Adjacent'' training scenario. For comparison, we also list the results in an In-Context scenario, where the required knowledge are provided in the prompt for each test case. 

As shown in Table~\ref{tab:mainResult}, neither the Separate nor the Adjacent training methods significantly outperform the random baseline in any dataset. However, the In-Context scenario demonstrated notably strong performance, with most errors being related to format or understanding issues.

This surprising result suggests that with only training on \(T_{1}\) and \(T_{2}\), the models struggle to effectively infer \(\bar{T}\), indicating a relatively weak OCKR capability. Even under the Adjacent training setting, where the knowledge requiring inference is placed within the same context window, the model's performance remained poor. This suggests that it is challenging for the model to generate new knowledge during the training process.

The results on Baichuan2-13B-CHAT, Pythia-12B, LLaMA2-7B-CHAT and LLaMA3-8B-Instruct (Appendix \ref{sec:otherModelResult}) are consistent with those of LLaMA2-13B-CHAT, showing that this is a common weakness among models with this size of parameters.

\begin{table*}[ht]
\centering
\footnotesize
\begin{tabular}{lrr}
\toprule
\textbf{Reasoning Pattern} & \textbf{Random} & \textbf{Reasoning Training} \\
\midrule
$A \land A\rightarrow R$ (simple) & 50.0 & 56 \\
$A \land R\rightarrow A$ (simple) & 5.0 & 7.5 \\
$R \land R\rightarrow R$ (simple) & 50.0 & 59.5 \\
\bottomrule
\end{tabular}
\caption{Performance in Scenarios with Reasoning Training: This table shows the impacts of employing reasoning training in three reasoning patterns. 
}
\label{tab:resultComplete}
\end{table*}

\subsection{Results with Reasoning Training}

We employed training with reasoning examples to enhance the model's reasoning capabilities. As depicted in Table \ref{tab:resultComplete}, across the three reasoning datasets, the model's performance is only slightly higher than the random baseline. Ten thousand instances are used for training these simple binary OCKR tasks. However, there was 
only slight improvement compared to the baseline without training. Thus, using reasoning data to improve the model's reasoning capabilities does not effectively enhance the model's OCKR abilities during the inference phase. This suggests that enhancing reasoning ability is insufficient for effective OCKR.

\subsection{Results with Retrieval Training}

\begin{table*}[ht]
\centering
\footnotesize
\begin{tabular}{lrrr}
\toprule
\textbf{Reasoning Pattern} & \textbf{Random} & \textbf{Retrieval Training} & \textbf{Retrieval Accuracy} \\
\midrule
$A \land A\rightarrow R$ (simple) & 50.0 & 93.5 & 89.8 \\
$A \land R\rightarrow A$ (simple) & 5.0 & 7.5 & 0.0 \\
$R \land R\rightarrow R$ (simple) & 50.0 & 52.0 & 0.0 \\
\bottomrule
\end{tabular}
\caption{Performance in Scenarios with Retrieval training: This table shows the impacts of employing Retrieval training in three reasoning patterns. The retrieval accuracy of required knowledge is also collected for each setting. 
}
\label{tab:Cot}
\end{table*}

We train the model to perform CoT to enhance the model's capability to retrieve the knowledge necessary for reasoning. Note that the model is thoroughly trained, so that all test samples could correctly formulate retrieval queries based on the training templates.
The results, as illustrated in Table \ref{tab:Cot}, show that the $A \land A \rightarrow R$ reasoning exhibits strong performance. This suggests that the main limitation of OCKR in such cases is that the model does not have the ability to automatically retrieval the related knowledge.

On contrary, both the $A \land R \rightarrow A$ and $R \land R \rightarrow R$ reasoning only surpass the random baseline by a small margin, showing extra difficulties in performing reasoning with relations. 

For further understanding of the problem, we evaluate the retrieval accuracy of the test examples \footnote{Please note that checking retrieval is only possible in this scenario, because there is an explicit process of knowledge retrieval.}. Results are listed in Table~\ref{tab:Cot}. When retrieving only attribute-type knowledge, as in the $A \land A \rightarrow R$ reasonings, the model performed well with 89.8\% accuracy. Thus it obtained more accurate answers (93.5\%). 
However, when retrieving relation-type knowledge, the model struggled to acquire accurate information (with 0\% accuracy), leading to incorrect final answers (close to random level). This indicates that even if we help the model determine the correct relation to retrieve (by explict training), it is still challenging to retrieve the second entity based on the first entity and the relation. This issue is closely related to the ``reversal curse'' (A is B cannot infer B is A) \cite{berglund2023reversal}, where models exhibit unidirectional learning limitations.

\subsection{Results of Cross-Lingual Reasoning}

We also analyze the cross-lingual OCKR capabilities as a special form of $A \land R\rightarrow A$. 
The results (presented Table \ref{tab:cross-lingual}) show that in cross-lingual scenarios, both the Separate and Adjacent training strategies outperform the random level, showing that a small portion of the knowledge could be inferred with another language, while the overall performance is still far from satisfaction (around 10\% in average). 

Comparing with the results in Table \ref{tab:mainResult}, it seems that the translation relation may exhibit a different learning mechanism from other relations. It might be a little easier to be extracted and utilized compared to the monolingual relations tested.

It is also easy to notice the diversity among languages, with German and Polish achieving the highest score compared to other test languages.

\begin{table}[ht]
\centering
\footnotesize
\begin{tabular}{lrr}
\toprule
\textbf{Language} & \textbf{Separate} & \textbf{Adjacent} \\
\midrule
de & 18.0 & 18.0 \\
zh & 8.5 & 11.0 \\
ar & 4.0 & 6.0 \\
he & 6.5 & 9.0 \\
ja & 7.0 & 9.0 \\
fr & 8.5 & 10.0 \\
it & 8.5 & 9.0 \\
pl & 16.5 & 18.0 \\
ru & 9.0 & 12.5 \\\hline
average & 9.6 & 11.4 \\
random & 5.0 & 5.0 \\ 
\bottomrule
\end{tabular}
\caption{Performance in Cross-Lingual OCKR Scenarios.
}
\label{tab:cross-lingual}
\end{table}

\section{Related Work}
\paragraph{Out-of-Context.} Krasheninnikov et al. \shortcite{krasheninnikov2023out} discuss how LLMs tend to internalize text that appears authentic or authoritative and apply it appropriately in context. Berglund et al. \shortcite{berglund2023taken} investigate LLM's situational awareness, particularly their ability to recognize their status as models and whether they are in a testing or deployment phase, 
proposing Out-of-Context Reasoning as an essential skill. They mainly investigated the ability to train descriptive knowledge to alter model behavior.
 In a different vein, Allen et al. \shortcite{allen2023physics} focus on LLM's ability to manipulate stored knowledge, especially in tasks like retrieval, classification, and comparison.
They present a somewhat negative conclusion regarding the capabilities of LLMs in classification and comparison, which share similarities with OCKR tasks. However, our approach differs significantly. Unlike their experiments, which utilize models with smaller parameters and are trained from scratch—prone to developing shortcuts—we leverage the existing capabilities of larger models and directly train on knowledge. 
 Additionally, Berglund et al.\shortcite{berglund2023reversal} highlight the ``Reversal Curse'' in LLMs, a limitation where models fail to generalize learned sentence structures to their reverse forms.

\paragraph{In-context.} Brown et al. \shortcite{brown2020language} introduce the concept of situational learning in LLMs, enabling them to leverage a few examples and pre-trained knowledge for improved task performance. Kojima et al. \shortcite{kojima2022large} explore LLM's zero-shot reasoning enhancement through task description integration, allowing models to utilize inherent knowledge for generalization. Wei et al. \shortcite{wei2022chain} demonstrate how LLMs can enhance complex reasoning with CoT prompting, crucial for intricate problem-solving.
Fang et al.\shortcite{fang2021boat} first defines the problem of inferring Concepts Out of the Dialogue Context in dialogue summarization. Hamilton et al. \shortcite{hamilton2018embedding} discusses how to effectively predict complex logical queries on incomplete knowledge graphs.

\paragraph{Cross-lingual.} Ye et al. \shortcite{ye2023language} present a comprehensive study comparing multilingual pre-trained models and English-centric models across various reasoning tasks. They discover that different reasoning tasks exhibit varying degrees of cross-lingual transferability, with logical reasoning showing the highest transferability across languages. Wang et al. \shortcite{wang2023seaeval} introduced SeaEval, a comprehensive benchmark designed to evaluate these models across a variety of aspects. Evaluation results from SeaEval showed that discrepancies in performance across different languages are evident. Qi et al. \shortcite{qi2023cross} propose a novel metric, Ranking-based Consistency, to evaluate the consistency of knowledge across languages independently from accuracy. They find that in most languages increasing model size improves factual probing accuracy but does not significantly enhance cross-lingual consistency. Gao et al. \shortcite{gao2024multilingual} constructed three types of testing datasets to evaluate cross-lingual knowledge alignment. Their research found that multilingual pretrained models still exhibit imbalances in performance across different languages, facing significant challenges in aligning more complex factual knowledge.

\section{Conclusion and Discussion}
This study comprehensively assesses the Out-of-Context Knowledge Reasoning capabilities of LLMs across various reasoning tasks. Our results show that the current LLMs are limited in performing out-of-context knowledge reasoning.

Through step-by-step experiments, we have identified two key reasons for the model's weak OCKR abilities: first, its difficulty in retrieving relational knowledge, and second, its inability to actively retrieve non-direct knowledge required for reasoning.


Firstly, the model struggles to retrieve relational knowledge, making it challenging to complete OCKR tasks. This issue is closely related to the ``reversal curse'' \cite{berglund2023reversal}, where models fail to generalize bidirectional relationships. To address this, methods such as Reverse Training \cite{berglund2023reversal}, Semantic-aware Permutation Training \cite{guo2024mitigating}, or Bidirectional Causal Language Modeling Optimization \cite{lv2023we} may be necessary to improve the model’s ability to retrieve relational knowledge.


Secondly, the model does not actively retrieve non-direct knowledge required for reasoning. We believe the limitation is related to the model's computational pathways capacity when predicting the next token. According to the Goyal et al. \shortcite{goyal2023think}, the computational pathways for obtaining each token result are limited, and retrieving non-directly related knowledge require much more pathways than retrieving directly related knowledge, which could not be obtained by current style of pretraining. 

To mitigate this issue, the model needs to identify when additional knowledge or planning is required during the output process. This may be achieved by inserting an appropriate number of Pause Tokens \citep{goyal2023think,herel2024thinking,zelikman2024quiet,wang2023guiding} when additional knowledge retrieval is necessary or by fine-tuning the model with step-by-step reasoning data for a specific task.

\section*{Limitations}
One major limitation of this study is that the evaluation is restricted to a few selected models, with the largest model being only 13B parameters. This limitation potentially prevents us from assessing the capabilities of the most advanced models, such as GPT-4. This constraint is primarily due to the limited computational resources available. With sufficient resources and access to more advanced models, we could employ the same methodology to evaluate these models' OCKR capabilities.

Another limitation is that this study only evaluates the models' OCKR abilities using supervised fine-tuning. It does not consider the impact of other training stages, such as reinforcement learning from human feedback \cite{zheng2023secrets}, on the models' OCKR abilities.

\section*{Ethics Statement}
The authors declare no competing interests. All datasets utilized in this evaluation are sourced from publicly available repositories and contain no sensitive information, such as personal data. Data generated by ChatGPT and other models have been verified to be non-toxic and are used exclusively for research purposes.

\section*{Acknowledgements}
We would like to thank the anonymous reviewers for their insightful comments. Shujian Huang is the corresponding author. This work is supported by National Science Foundation of China (No. 62376116, 62176120) and Nanjing University-China Mobile Communications Group Co.,Ltd. Joint Institute.

\appendix
\renewcommand{\thetable}{A\arabic{table}}
\setcounter{table}{0} 

\begin{table*}[!ht]
\centering
\footnotesize
\begin{tabular}{p{0.3\linewidth}p{0.65\linewidth}}
\toprule
\textbf{Knowledge Template} & \textbf{Data Example} \\
\midrule
(x,birth\_year\_equals,y) & 
Q: Did XenoGlimmer and MeadowGlitter share the same birth year? \newline
A: Yes, MeadowGlitter and XenoGlimmer were born in the same year. \\
\hline
(x,not\_birth\_year\_equals,y$'$) & 
Q: Did InfiniteBreeze and XenoGlimmer share the same birth year? \newline
A: No, InfiniteBreeze and XenoGlimmer were not born in the same year. \\
\hline
(y,birth\_year,year) & 
Q: In what year was XenoGlimmer born? \newline
A: XenoGlimmer was born in the year 2010. \\
\hline
(y,birth\_year\_greater\_than,z$_{\text{small}}$) & 
Q: Does GlacialHarmony have more years of life than MeadowGlitter? \newline
A: Yes, GlacialHarmony does have more years of life than MeadowGlitter. \\
\hline
(y,not\_birth\_year\_greater\_than,z$_{\text{large}}$) & 
Q: Does GlacialHarmony have more years of life than InfiniteMeadow? \newline
A: No, GlacialHarmony does not have more years of life than InfiniteMeadow. \\
\hline
(y,one\_year\_older\_than,z) & 
Q: Could you confirm if UnseenMeadow was born a year earlier than FieryCascade? \newline
A: Yes, it is confirmed that UnseenMeadow was born one year before FieryCascade. \\
\hline
(y,not\_birth\_year\_greater\_than\_1,z$'$) & 
Q: Could you confirm if UnseenMeadow was born a year earlier than StellarPulse? \newline
A: No, it is not true that UnseenMeadow was born a year before StellarPulse. \\
\hline
(x,parents\_generation,y) & 
Q: Is the parents' generation of XenoGlimmer EclipseQuiver? \newline
A: Yes, the parents' generation of XenoGlimmer is EclipseQuiver. \\
\hline
(x,not\_parents\_generation,y$'$) & 
Q: Is the parents' generation of IrisWander EclipseQuiver? \newline
A: No, the parents' generation of IrisWander is not EclipseQuiver. \\
\hline
(x,grandparents\_generation,z) & 
Q: Is the grandparents' generation of XenoGlimmer MeadowGlitter? \newline
A: Yes, the grandparents' generation of XenoGlimmer is MeadowGlitter. \\
\hline
(x,not\_grandparents\_generation,z$'$) &
Q: Is the grandparents' generation of XenoGlimmer IridescentDream? \newline
A: No, the grandparents' generation of XenoGlimmer is not IridescentDream. \\
\hline
(x$_{\text{de}}$,birth\_year,year) &
Q: In welchem Jahr wurde XenoSchimmer geboren? \newline
A: XenoSchimmer wurde im Jahr 2010 geboren. \\
\hline
(x$_{\text{en}}$,translation,x$_{\text{de}}$) & 
Q: Could you convert the upcoming English text to German? \newline
Input: XenoGlimmer \newline
A: XenoSchimmer \\
\hline
Adjacent \newline
(x,birth\_year\_equals,y) \newline
(x,birth\_year\_equals,y) & 
Q: Did EclipseQuiver and XenoGlimmer share the same birth year? Did MeadowGlitter and XenoGlimmer share the same birth year? \newline
A: Yes, EclipseQuiver and XenoGlimmer were born in the same year. Yes, MeadowGlitter and XenoGlimmer were born in the same year. \\
\hline
Adjacent \newline
(x$_{\text{en}}$,birth\_year,year) \newline
(x$_{\text{en}}$,translation,x$_{\text{de}}$) & 
Q: Can you tell me the birth year of MysticDawn (German: MystischerMorgen)? \newline
A: The birth year of MysticDawn (German: MystischerMorgen) is 1992. \\
\bottomrule
\end{tabular}
\caption{Illustrative Examples of Data in the ID-OCKR dataset for different knowledge templates in separate and adjacent training scenarios}
\label{tab:knowledge_triples}
\end{table*}

\section{Data Sample}
\label{sec:DataSample}
The actual training data examples corresponding to the knowledge triples in the article can be seen in Table \ref{tab:knowledge_triples}.

\section{Additional detailed description of the dataset}
\label{sec:detailDataset}
In this section, we introduce additional details on how the dataset is processed.

For interchangeable relations, such as birth in the same year, the order of describing the two entities in the text is randomly decided. For other relations, training and testing text have the same order of mentioning the two entities, to avoid the reversal curse \cite{berglund2023reversal}. 

In the $A \land A \rightarrow R$ (hard) dataset, due to the presence of the largest birth year, the individual with the latest birth year is excluded from comparisons. In the $A \land A \rightarrow R$ and Cross-Lingual Reasoning datasets, the lack of training templates corresponding to the test templates makes accurate testing challenging. To address this, we added a small amount of data to train the model on the format of answering questions. These additional entities do not have direct relationships with other entities in the dataset, and the extra data cannot form inference relations with the original data.

\section{Experiments Details}
\label{sec:Hyper-parameters}
This section outlines the details of our experiments for reproducibility.

\subsection{Used Scientific Artifacts}
We used the following scientific artifacts in our research:
\begin{itemize}
\item \textit{PyTorch}~(\citealp{Ansel_PyTorch_2_Faster_2024}, BSD license), a framework for building and running deep learning models.
\item \textit{Transformers}~(\citealp{wolf-etal-2020-transformers}, Apache-2.0 license), a library providing a user friendly interface for running and fine-tuning pre-trained models.
\item \textit{DeepSpeed}~(\citealp{rasley2020deepspeed}, Apache-2.0 license), a library optimizing the parallel training of the deep learning models.
\item \textit{LLaMA-Factory}~(\citealp{zheng2024llamafactory}, Apache-2.0 license), a library that provides a unifying way to easily fine-tune large language models with parameter efficient fine-tuning technique like LoRA.
\end{itemize}
\subsection{Hyperparameters}
For model inference, the temperature parameter is set to 0. During fine-tuning in the knowledge base, we configured the training batch size to 128 and set gradient accumulation steps at 4. The maximum number of steps is limited to 300. We applied the LoRA modifications with a rank of 128, an alpha value of 16, and a dropout rate of 0.05. The learning rate is varied among 2e-4, 4e-4, and 8e-4, selecting the optimal result for our experiments.

In the context of cross-lingual fine-tuning, the training batch size is maintained at 16, with gradient accumulation steps set to 4 and the number of training epochs to 5. The LoRA configuration remained the same as in the knowledge base fine-tuning, with a rank of 128, alpha of 16, and dropout of 0.05. The learning rate for these experiments is set to 2e-4.

\subsection{Computation resources}
Our computational resources were limited to V100 GPUs, allowing us to fine-tune 13B models with LoRA or fully fine-tune 7B models.


\section{Validation of Results with Additional Models}
\label{sec:otherModelResult}

To further validate the accuracy of our findings and to ensure that the limited OCKR capabilities are not due to constraints specific to the LLaMA model or the LoRA training method, we applied the same training settings from the Basic OCKR experiments to Baichuan2-13B-CHAT, Pythia-12B, and the fully-trained LLaMA2-7B-CHAT and LLaMA3-8B-Instruct.

The experimental results are presented in Tables \ref{tab:baichuanResult}, \ref{tab:pythiaResult}, \ref{tab:FullResult} and \ref{tab:llama3} respectively. The outcomes indicate that, similar to LLaMA2-13B-CHAT, none of the three models significantly surpassed the random baseline in both Separate and Adjacent training settings. These consistent findings suggest the inherent limitations of the current models in achieving robust OCKR capabilities.

\begin{table}[!ht]
\centering
\footnotesize
\begin{tabular}{lrrr}
\toprule
\textbf{Dataset} & \textbf{Random} & \textbf{Separate} & \textbf{Adjacent} \\
\midrule
$A \land A\rightarrow R$ (simple) & 50.0 & 50.5 & 50.0 \\
\addlinespace
$A \land R\rightarrow A$ (simple) & 5.0 & 4.5 & 7.5 \\
\addlinespace
$R \land R\rightarrow R$ (simple) & 50.0 & 51.5 & 50.25 \\
\addlinespace
$A \land A\rightarrow R$ (hard) & 50.0 & 54.7 & 52.6 \\
\addlinespace
$A \land R\rightarrow A$ (hard) & 5.0 & 6.5 & 6.0 \\
\addlinespace
$R \land R\rightarrow R$ (hard) & 50.0 & 50.25 & 50.75 \\
\bottomrule
\end{tabular}
\caption{Basic OCKR experiment results for the Baichuan2-13B-CHAT model.}
\label{tab:baichuanResult}
\end{table}

\begin{table}[!ht]
\centering
\footnotesize
\begin{tabular}{lrrr}
\toprule
\textbf{Dataset} & \textbf{Random} & \textbf{Separate} & \textbf{Adjacent} \\
\midrule
$A \land A\rightarrow R$ (simple) & 50.0 & 50.75 & 53.25 \\
\addlinespace
$A \land R\rightarrow A$ (simple) & 5.0 & 5.5 & 7.5 \\
\addlinespace
$R \land R\rightarrow R$ (simple) & 50.0 & 50.5 & 52.25 \\
\addlinespace
$A \land A\rightarrow R$ (hard) & 50.0 & 56.8 & 59.7 \\
\addlinespace
$A \land R\rightarrow A$ (hard) & 5.0 & 6.0 & 7.5 \\
\addlinespace
$R \land R\rightarrow R$ (hard) & 50.0 & 50.75 & 50.5 \\
\bottomrule
\end{tabular}
\caption{Basic OCKR experiment results for the Pythia-12B model.}
\label{tab:pythiaResult}
\end{table}

\begin{table}[!ht]
\centering
\footnotesize
\begin{tabular}{lrrr}
\toprule
\textbf{Dataset} & \textbf{Random} & \textbf{Separate} & \textbf{Adjacent} \\
\midrule
$A \land A\rightarrow R$ (simple) & 50.0 & 51.0 & 49.0 \\
\addlinespace
$A \land R\rightarrow A$ (simple) & 5.0 & 5.5 & 5.5 \\
\addlinespace
$R \land R\rightarrow R$ (simple) & 50.0 & 52.5 & 50.25 \\
\addlinespace
$A \land A\rightarrow R$ (hard) & 50.0 & 50.0 & 54.47 \\
\addlinespace
$A \land R\rightarrow A$ (hard) & 5.0 & 6.0 & 5.5 \\
\addlinespace
$R \land R\rightarrow R$ (hard) & 50.0 & 49.5 & 52.75 \\
\bottomrule
\end{tabular}
\caption{Basic OCKR experiment results for the LLaMA2-7B-CHAT model.}
\label{tab:FullResult}
\end{table}

\begin{table}[!ht]
\centering
\footnotesize
\begin{tabular}{lrrr}
\toprule
\textbf{Dataset} & \textbf{Random} & \textbf{Separate} & \textbf{Adjacent} \\
\midrule
$A \land A\rightarrow R$ (simple) & 50.0 & 52.25 & 49.3 \\
\addlinespace
$A \land R\rightarrow A$ (simple) & 5.0 & 8.0 & 3.5 \\
\addlinespace
$R \land R\rightarrow R$ (simple) & 50.0 & 50.8 & 50.8 \\
\addlinespace
$A \land A\rightarrow R$ (hard) & 50.0 & 53.5 & 53.0 \\
\addlinespace
$A \land R\rightarrow A$ (hard) & 5.0 & 7.5 & 6.0 \\
\addlinespace
$R \land R\rightarrow R$ (hard) & 50.0 & 49.3 & 49.8 \\
\bottomrule
\end{tabular}
\caption{Basic OCKR experiment results for the LLaMA3-8B-Instruct model.}
\label{tab:llama3}
\end{table}


\begin{thebibliography}{30}
\providecommand{\natexlab}[1]{#1}
\bibitem[{Goyal et~al.(2023)Goyal, Ji, Rawat, Menon, Kumar, and Nagarajan}]{goyal2023think}
Sachin Goyal, Ziwei Ji, Ankit Singh Rawat, Aditya Krishna Menon, Sanjiv Kumar, and Vaishnavh Nagarajan. 2023.
\newblock Think before you speak: Training language models with pause tokens.
\newblock \emph{arXiv preprint arXiv:2310.02226}.
\bibitem[{Herel and Mikolov(2024)}]{herel2024thinking}
David Herel and Tomas Mikolov. 2024.
\newblock Thinking tokens for language modeling.
\newblock \emph{arXiv preprint arXiv:2405.08644}.

\bibitem[{Guo et~al.(2024)Guo, Wang, Guo, Tan, Bian, and Yang}]{guo2024mitigating}
Qingyan Guo, Rui Wang, Junliang Guo, Xu Tan, Jiang Bian, and Yujiu Yang. 2024.
\newblock Mitigating reversal curse via semantic-aware permutation training.
\newblock \emph{arXiv preprint arXiv:2403.00758}.

\bibitem[{Lv et~al.(2023)Lv, Zhang, Xie, Tu, Chen, Wen, and Yan}]{lv2023we}
Ang Lv, Kaiyi Zhang, Shufang Xie, Quan Tu, Yuhan Chen, Ji-Rong Wen, and Rui Yan. 2023.
\newblock Are we falling in a middle-intelligence trap? An analysis and mitigation of the reversal curse.
\newblock \emph{arXiv preprint arXiv:2311.07468}.


\bibitem[{Zelikman et~al.(2024)Zelikman, Harik, Shao, Jayasiri, Haber, and Goodman}]{zelikman2024quiet}
Eric Zelikman, Georges Harik, Yijia Shao, Varuna Jayasiri, Nick Haber, and Noah~D. Goodman. 2024.
\newblock Quiet-star: Language models can teach themselves to think before speaking.
\newblock \emph{arXiv preprint arXiv:2403.09629}.

\bibitem[{Wang et~al.(2023)Wang, Caccia, Ostapenko, Yuan, and Sordoni}]{wang2023guiding}
Xinyi Wang, Lucas Caccia, Oleksiy Ostapenko, Xingdi Yuan, and Alessandro Sordoni. 2023.
\newblock Guiding language model reasoning with planning tokens.
\newblock \emph{arXiv preprint arXiv:2310.05707}.
\bibitem[{Achiam et~al.(2023)Achiam, Adler, Agarwal, Ahmad, Akkaya, Aleman, Almeida, Altenschmidt, Altman, Anadkat et~al.}]{achiam2023gpt}
Josh Achiam, Steven Adler, Sandhini Agarwal, Lama Ahmad, Ilge Akkaya, Florencia~Leoni Aleman, Diogo Almeida, Janko Altenschmidt, Sam Altman, Shyamal Anadkat, et~al. 2023.
\newblock Gpt-4 technical report.
\newblock \emph{arXiv preprint arXiv:2303.08774}.

\bibitem[{AlKhamissi et~al.(2022)AlKhamissi, Li, Celikyilmaz, Diab, and Ghazvininejad}]{alkhamissi2022review}
Badr AlKhamissi, Millicent Li, Asli Celikyilmaz, Mona Diab, and Marjan Ghazvininejad. 2022.
\newblock A review on language models as knowledge bases.
\newblock \emph{arXiv preprint arXiv:2204.06031}.

\bibitem[{Allen-Zhu and Li(2023)}]{allen2023physics}
Zeyuan Allen-Zhu and Yuanzhi Li. 2023.
\newblock Physics of language models: Part 3.2, knowledge manipulation.
\newblock \emph{arXiv preprint arXiv:2309.14402}.

\bibitem[{Ansel et~al.(2024)Ansel, Yang, He, Gimelshein, Jain, Voznesensky, Bao, Bell, Berard, Burovski, Chauhan, Chourdia, Constable, Desmaison, DeVito, Ellison, Feng, Gong, Gschwind, Hirsh, Huang, Kalambarkar, Kirsch, Lazos, Lezcano, Liang, Liang, Lu, Luk, Maher, Pan, Puhrsch, Reso, Saroufim, Siraichi, Suk, Suo, Tillet, Wang, Wang, Wen, Zhang, Zhao, Zhou, Zou, Mathews, Chanan, Wu, and Chintala}]{Ansel_PyTorch_2_Faster_2024}
Jason Ansel, Edward Yang, Horace He, Natalia Gimelshein, Animesh Jain, Michael Voznesensky, Bin Bao, Peter Bell, David Berard, Evgeni Burovski, Geeta Chauhan, Anjali Chourdia, Will Constable, Alban Desmaison, Zachary DeVito, Elias Ellison, Will Feng, Jiong Gong, Michael Gschwind, Brian Hirsh, Sherlock Huang, Kshiteej Kalambarkar, Laurent Kirsch, Michael Lazos, Mario Lezcano, Yanbo Liang, Jason Liang, Yinghai Lu, CK~Luk, Bert Maher, Yunjie Pan, Christian Puhrsch, Matthias Reso, Mark Saroufim, Marcos~Yukio Siraichi, Helen Suk, Michael Suo, Phil Tillet, Eikan Wang, Xiaodong Wang, William Wen, Shunting Zhang, Xu~Zhao, Keren Zhou, Richard Zou, Ajit Mathews, Gregory Chanan, Peng Wu, and Soumith Chintala. 2024.
\newblock \href {https://doi.org/10.1145/3620665.3640366} {{PyTorch 2: Faster Machine Learning Through Dynamic Python Bytecode Transformation and Graph Compilation}}.
\newblock In \emph{29th ACM International Conference on Architectural Support for Programming Languages and Operating Systems, Volume 2 (ASPLOS '24)}. ACM.

\bibitem[{Berglund et~al.(2023{\natexlab{a}})Berglund, Stickland, Balesni, Kaufmann, Tong, Korbak, Kokotajlo, and Evans}]{berglund2023taken}
Lukas Berglund, Asa~Cooper Stickland, Mikita Balesni, Max Kaufmann, Meg Tong, Tomasz Korbak, Daniel Kokotajlo, and Owain Evans. 2023{\natexlab{a}}.
\newblock Taken out of context: On measuring situational awareness in llms.
\newblock \emph{arXiv preprint arXiv:2309.00667}.

\bibitem[{Berglund et~al.(2023{\natexlab{b}})Berglund, Tong, Kaufmann, Balesni, Stickland, Korbak, and Evans}]{berglund2023reversal}
Lukas Berglund, Meg Tong, Max Kaufmann, Mikita Balesni, Asa~Cooper Stickland, Tomasz Korbak, and Owain Evans. 2023{\natexlab{b}}.
\newblock The reversal curse: Llms trained on`` a is b'' fail to learn`` b is a''.
\newblock \emph{arXiv preprint arXiv:2309.12288}.

\bibitem[{Besta et~al.(2023)Besta, Blach, Kubicek, Gerstenberger, Gianinazzi, Gajda, Lehmann, Podstawski, Niewiadomski, Nyczyk et~al.}]{besta2023graph}
Maciej Besta, Nils Blach, Ales Kubicek, Robert Gerstenberger, Lukas Gianinazzi, Joanna Gajda, Tomasz Lehmann, Michal Podstawski, Hubert Niewiadomski, Piotr Nyczyk, et~al. 2023.
\newblock Graph of thoughts: Solving elaborate problems with large language models.
\newblock \emph{arXiv preprint arXiv:2308.09687}.

\bibitem[{Biderman et~al.(2023)Biderman, Schoelkopf, Anthony, Bradley, O’Brien, Hallahan, Khan, Purohit, Prashanth, Raff et~al.}]{biderman2023pythia}
Stella Biderman, Hailey Schoelkopf, Quentin~Gregory Anthony, Herbie Bradley, Kyle O’Brien, Eric Hallahan, Mohammad~Aflah Khan, Shivanshu Purohit, USVSN~Sai Prashanth, Edward Raff, et~al. 2023.
\newblock Pythia: A suite for analyzing large language models across training and scaling.
\newblock In \emph{International Conference on Machine Learning}, pages 2397--2430. PMLR.

\bibitem[{Brown et~al.(2020)Brown, Mann, Ryder, Subbiah, Kaplan, Dhariwal, Neelakantan, Shyam, Sastry, Askell et~al.}]{brown2020language}
Tom Brown, Benjamin Mann, Nick Ryder, Melanie Subbiah, Jared~D Kaplan, Prafulla Dhariwal, Arvind Neelakantan, Pranav Shyam, Girish Sastry, Amanda Askell, et~al. 2020.
\newblock Language models are few-shot learners.
\newblock \emph{Advances in neural information processing systems}, 33:1877--1901.

\bibitem[{Fang et~al.(2021)Fang, Pan, Zhang, Song, Xu, and Yu}]{fang2021boat}
Tianqing Fang, Haojie Pan, Hongming Zhang, Yangqiu Song, Kun Xu, and Dong Yu. 2021.
\newblock Do boat and ocean suggest beach? dialogue summarization with external knowledge.
\newblock In \emph{3rd Conference on Automated Knowledge Base Construction}.

\bibitem[{Gao et~al.(2024)Gao, Hu, Hu, Chen, Li, and Huang}]{gao2024multilingual}
Changjiang Gao, Hongda Hu, Peng Hu, Jiajun Chen, Jixing Li, and Shujian Huang. 2024.
\newblock Multilingual pretraining and instruction tuning improve cross-lingual knowledge alignment, but only shallowly.
\newblock \emph{arXiv preprint arXiv:2404.04659}.

\bibitem[{Hamilton et~al.(2018)Hamilton, Bajaj, Zitnik, Jurafsky, and Leskovec}]{hamilton2018embedding}
Will Hamilton, Payal Bajaj, Marinka Zitnik, Dan Jurafsky, and Jure Leskovec. 2018.
\newblock Embedding logical queries on knowledge graphs.
\newblock \emph{Advances in neural information processing systems}, 31.

\bibitem[{Ho et~al.(2022)Ho, Schmid, and Yun}]{ho2022large}
Namgyu Ho, Laura Schmid, and Se-Young Yun. 2022.
\newblock Large language models are reasoning teachers.
\newblock \emph{arXiv preprint arXiv:2212.10071}.

\bibitem[{Hu et~al.(2021)Hu, Shen, Wallis, Allen-Zhu, Li, Wang, Wang, and Chen}]{hu2021lora}
Edward~J Hu, Yelong Shen, Phillip Wallis, Zeyuan Allen-Zhu, Yuanzhi Li, Shean Wang, Lu~Wang, and Weizhu Chen. 2021.
\newblock Lora: Low-rank adaptation of large language models.
\newblock \emph{arXiv preprint arXiv:2106.09685}.

\bibitem[{Kejriwal et~al.(2021)Kejriwal, Knoblock, and Szekely}]{kejriwal2021knowledge}
Mayank Kejriwal, Craig~A Knoblock, and Pedro Szekely. 2021.
\newblock \emph{Knowledge graphs: Fundamentals, techniques, and applications}.
\newblock MIT Press.

\bibitem[{Kojima et~al.(2022)Kojima, Gu, Reid, Matsuo, and Iwasawa}]{kojima2022large}
Takeshi Kojima, Shixiang~Shane Gu, Machel Reid, Yutaka Matsuo, and Yusuke Iwasawa. 2022.
\newblock Large language models are zero-shot reasoners.
\newblock \emph{Advances in neural information processing systems}, 35:22199--22213.

\bibitem[{Krasheninnikov et~al.(2023)Krasheninnikov, Krasheninnikov, and Krueger}]{krasheninnikov2023out}
Dmitrii Krasheninnikov, Egor Krasheninnikov, and David Krueger. 2023.
\newblock Out-of-context meta-learning in large language models.
\newblock In \emph{ICLR 2023 Workshop on Mathematical and Empirical Understanding of Foundation Models}.

\bibitem[{Petroni et~al.(2019)Petroni, Rockt{\"a}schel, Lewis, Bakhtin, Wu, Miller, and Riedel}]{petroni2019language}
Fabio Petroni, Tim Rockt{\"a}schel, Patrick Lewis, Anton Bakhtin, Yuxiang Wu, Alexander~H Miller, and Sebastian Riedel. 2019.
\newblock Language models as knowledge bases?
\newblock \emph{arXiv preprint arXiv:1909.01066}.

\bibitem[{Qi et~al.(2023)Qi, Fern{\'a}ndez, and Bisazza}]{qi2023cross}
Jirui Qi, Raquel Fern{\'a}ndez, and Arianna Bisazza. 2023.
\newblock Cross-lingual consistency of factual knowledge in multilingual language models.
\newblock \emph{arXiv preprint arXiv:2310.10378}.

\bibitem[{Rasley et~al.(2020)Rasley, Rajbhandari, Ruwase, and He}]{rasley2020deepspeed}
Jeff Rasley, Samyam Rajbhandari, Olatunji Ruwase, and Yuxiong He. 2020.
\newblock Deepspeed: System optimizations enable training deep learning models with over 100 billion parameters.
\newblock In \emph{Proceedings of the 26th ACM SIGKDD International Conference on Knowledge Discovery \& Data Mining}, pages 3505--3506.

\bibitem[{Touvron et~al.(2023)Touvron, Martin, Stone, Albert, Almahairi, Babaei, Bashlykov, Batra, Bhargava, Bhosale et~al.}]{touvron2023llama}
Hugo Touvron, Louis Martin, Kevin Stone, Peter Albert, Amjad Almahairi, Yasmine Babaei, Nikolay Bashlykov, Soumya Batra, Prajjwal Bhargava, Shruti Bhosale, et~al. 2023.
\newblock Llama 2: Open foundation and fine-tuned chat models.
\newblock \emph{arXiv preprint arXiv:2307.09288}.

\bibitem[{Wang et~al.(2023)Wang, Liu, Huang, Jiao, Ding, Aw, and Chen}]{wang2023seaeval}
Bin Wang, Zhengyuan Liu, Xin Huang, Fangkai Jiao, Yang Ding, Ai~Ti Aw, and Nancy~F Chen. 2023.
\newblock Seaeval for multilingual foundation models: From cross-lingual alignment to cultural reasoning.
\newblock \emph{arXiv preprint arXiv:2309.04766}.

\bibitem[{Wei et~al.(2022)Wei, Wang, Schuurmans, Bosma, Xia, Chi, Le, Zhou et~al.}]{wei2022chain}
Jason Wei, Xuezhi Wang, Dale Schuurmans, Maarten Bosma, Fei Xia, Ed~Chi, Quoc~V Le, Denny Zhou, et~al. 2022.
\newblock Chain-of-thought prompting elicits reasoning in large language models.
\newblock \emph{Advances in Neural Information Processing Systems}, 35:24824--24837.

\bibitem[{Wei et~al.(2023)Wei, Huang, Zhang, and Kwok}]{wei2023kicgpt}
Yanbin Wei, Qiushi Huang, Yu~Zhang, and James Kwok. 2023.
\newblock Kicgpt: Large language model with knowledge in context for knowledge graph completion.
\newblock In \emph{Findings of the Association for Computational Linguistics: EMNLP 2023}, pages 8667--8683.

\bibitem[{Wolf et~al.(2020)Wolf, Debut, Sanh, Chaumond, Delangue, Moi, Cistac, Rault, Louf, Funtowicz, Davison, Shleifer, von Platen, Ma, Jernite, Plu, Xu, Scao, Gugger, Drame, Lhoest, and Rush}]{wolf-etal-2020-transformers}
Thomas Wolf, Lysandre Debut, Victor Sanh, Julien Chaumond, Clement Delangue, Anthony Moi, Pierric Cistac, Tim Rault, Rémi Louf, Morgan Funtowicz, Joe Davison, Sam Shleifer, Patrick von Platen, Clara Ma, Yacine Jernite, Julien Plu, Canwen Xu, Teven~Le Scao, Sylvain Gugger, Mariama Drame, Quentin Lhoest, and Alexander~M. Rush. 2020.
\newblock \href {https://www.aclweb.org/anthology/2020.emnlp-demos.6} {Transformers: State-of-the-art natural language processing}.
\newblock In \emph{Proceedings of the 2020 Conference on Empirical Methods in Natural Language Processing: System Demonstrations}, pages 38--45, Online. Association for Computational Linguistics.

\bibitem[{Yang et~al.(2023)Yang, Xiao, Wang, Zhang, Bian, Yin, Lv, Pan, Wang, Yan et~al.}]{yang2023baichuan}
Aiyuan Yang, Bin Xiao, Bingning Wang, Borong Zhang, Ce~Bian, Chao Yin, Chenxu Lv, Da~Pan, Dian Wang, Dong Yan, et~al. 2023.
\newblock Baichuan 2: Open large-scale language models.
\newblock \emph{arXiv preprint arXiv:2309.10305}.

\bibitem[{Yao et~al.(2023)Yao, Yu, Zhao, Shafran, Griffiths, Cao, and Narasimhan}]{yao2023tree}
Shunyu Yao, Dian Yu, Jeffrey Zhao, Izhak Shafran, Thomas~L Griffiths, Yuan Cao, and Karthik Narasimhan. 2023.
\newblock Tree of thoughts: Deliberate problem solving with large language models.
\newblock \emph{arXiv preprint arXiv:2305.10601}.

\bibitem[{Ye et~al.(2023)Ye, Tao, and Kong}]{ye2023language}
Jiacheng Ye, Xijia Tao, and Lingpeng Kong. 2023.
\newblock Language versatilists vs. specialists: An empirical revisiting on multilingual transfer ability.
\newblock \emph{arXiv preprint arXiv:2306.06688}.

\bibitem[{Zheng et~al.(2023)Zheng, Dou, Gao, Hua, Shen, Wang, Liu, Jin, Liu, Zhou et~al.}]{zheng2023secrets}
Rui Zheng, Shihan Dou, Songyang Gao, Yuan Hua, Wei Shen, Binghai Wang, Yan Liu, Senjie Jin, Qin Liu, Yuhao Zhou, et~al. 2023.
\newblock Secrets of rlhf in large language models part i: Ppo.
\newblock \emph{arXiv preprint arXiv:2307.04964}.

\bibitem[{Zheng et~al.(2024)Zheng, Zhang, Zhang, Ye, Luo, and Ma}]{zheng2024llamafactory}
Yaowei Zheng, Richong Zhang, Junhao Zhang, Yanhan Ye, Zheyan Luo, and Yongqiang Ma. 2024.
\newblock \href {http://arxiv.org/abs/2403.13372} {Llamafactory: Unified efficient fine-tuning of 100+ language models}.
\newblock \emph{arXiv preprint arXiv:2403.13372}.

\end{thebibliography}
\end{document}